\documentclass[11pt]{article}

\usepackage[utf8]{inputenc}
\usepackage[T1]{fontenc}
\usepackage[margin=1in]{geometry}
\usepackage{amsmath,amssymb}
\usepackage{graphicx}
\usepackage{booktabs}
\usepackage{array}
\usepackage{siunitx}
\usepackage[hidelinks]{hyperref}
\usepackage{caption}
\usepackage{authblk}
\usepackage[round,authoryear]{natbib}
\usepackage{microtype}

\graphicspath{{figures/}}

\newcommand{\mnus}{\ensuremath{-}}

\title{\bfseries The Learning Objective Governs Perceptual Narrowing:\\
A Cross-Lingual, Layer-Wise, Ten-Seed Study of\\
Self-Supervised Speech Encoders}
\author[1]{Sejin Yoo}
\affil[1]{Independent Researcher\\
\texttt{loomy.sjyoo@gmail.com}}
\date{August 2026}

\begin{document}
\maketitle
\sloppy

\begin{abstract}
\noindent
Perceptual narrowing---the developmental loss of non-native phoneme discrimination in the first
year of life \citep{werker1984}---is a canonical developmental finding, yet \emph{what learning
objective produces it} remains open. We train a \(\sim\)7\,M-parameter Transformer encoder on
child-directed and read speech and evaluate phoneme ABX in English, French, and Mandarin over ten
seeds, the seed as the unit of replication. Six results. \textbf{(1)}~The objective sets the
direction of cross-lingual transfer: reconstruction (masked mel-prediction) degrades non-native
discrimination, prediction (frame-contrastive) improves it---a same-encoder, same-data gap of
\(+0.051\) in first-layer Mandarin ABX (\(p=3\times10^{-8}\)), unanimous in sign across twenty runs.
\textbf{(2)}~That decline combines a large arm-intrinsic difficulty gradient with a smaller
language-specialization effect (matched vs.\ mismatched \(+0.022\), \(p=10^{-4}\), all four layers).
\textbf{(3)}~Against a language-symmetric raw-mel floor, reconstruction pushes the first layer
\emph{below} the discriminability of its input; prediction pushes it \emph{above}.
\textbf{(4)}~Read speech gives a \(3.6\times\) steeper non-native decline than child-directed
speech. \textbf{(5)}~The customary three-seed budget cannot see this reliably: an effect
unambiguous at ten seeds is called significant by as few as 70\% of three-seed subsets.
\textbf{(6)}~Six objective configurations---sharpening, compression, consolidation, their
composition, and word-level semantic grounding in two forms---fail to produce the full developmental
signature (native improves \emph{and} non-native declines): a single objective moves both languages
the same way because it acts on a shared representation. We conclude that the objective, not the
architecture, is the first-order determinant of narrowing-shaped representational change.
\end{abstract}

\section{Introduction}

By 6--12 months, infants lose the ability to discriminate phoneme contrasts absent from their
native language while retaining native contrasts \citep{werker1984,kuhl1992}. This perceptual
narrowing is a canonical developmental-neuroscience finding, and it poses a computational question
that its neural description does not answer: \textbf{what learning objective, operating on early
speech input, produces experience-driven loss of non-native discrimination?}

Self-supervised learning (SSL) offers a controlled setting to ask it: a model with an explicit
objective, trained on speech, whose internal discriminability can be read out at every layer. The
recent computational language-acquisition literature has largely concluded that SSL speech models
reproduce the \emph{native-language advantage} but not the developmental \emph{decline}
\citep{schatz2021,rasanen2026,lavechin2025}. We revisit that conclusion and find it under-determined: the
answer depends on measurement choices---the layer, the contrast set, the training language, the
register, the seed count---that prior work has not varied together, and on the \emph{objective},
which the literature treats as a fixed background rather than the variable of interest.

Our central result is that the objective is the variable of interest. A reconstruction objective
and a prediction objective, holding the encoder, data, and seed identical, drive cross-lingual
transfer in opposite directions. We dissect the reconstruction-driven decline into two mechanisms,
ground it in an absolute representational reference (the input-feature floor), show its
register-sensitivity, and quantify how badly the field's seed budget under-powers it. We test a
brain-circuit-mapped architecture and find the objective, not the architecture, carries the
effect. Finally we search directly for the full signature with six objective configurations and
map why none reproduces it.

\paragraph{Contributions.}
(1)~The learning objective sets the direction of cross-lingual transfer (ten seeds, unanimous).
(2)~A training-language crossover separating arm-intrinsic difficulty from language specialization,
with a contrast-level confirmation.
(3)~A representation-geometry probe placing reconstruction below and prediction above the
input-feature floor.
(4)~A register control and a seed-budget fragility analysis, both methodological cautions with
quantified effect.
(5)~A brain-circuit dual-code region probe locating the negative architectural result.
(6)~A divergence search: six objective configurations engineered to reproduce the full signature,
a mechanism map showing why none does, and the selectivity requirement it implies.

\section{Background}

\paragraph{Perceptual narrowing and its neural description.}
Infants discriminate non-native contrasts at birth and lose this by 10--12 months for contrasts
absent from the native language, tested on exactly those contrasts (Hindi retroflex for English
learners; Mandarin tone for English learners) \citep{werker1984}. The Native Language Magnet
theory \citep{kuhl1992} frames it as experience warping perceptual space. The neural substrate is
described by dual-stream models of speech \citep{hickok2007} and, in the verbal-repetition circuit
this project departs from, by a dual code---an articulatory representation (left inferior frontal
gyrus, LIFG) and an acoustic-phonetic representation (left middle temporal gyrus, MTG)
\citep{yoo2013}. What none of these specify is the \emph{learning objective} whose optimization
produces the narrowing.

\paragraph{Self-supervised objectives.}
Two objective families dominate. \emph{Reconstruction} (masked-prediction, wav2vec2-style;
\citealp{baevski2020}) predicts masked input detail from context, optimizing signal fidelity.
\emph{Prediction} (contrastive predictive coding, CPC; \citealp{oord2018}) discriminates future
frames via InfoNCE, optimizing invariant structure. \citet{rasanen2026} reviews their
infant-learning track record: the native advantage is reproduced, the decline is not. The review
treats the objective as fixed; we vary it. Standard evaluation is final-layer ABX, over all
non-native contrasts, under a single objective, at a small seed count; Sections~\ref{sec:layer},
\ref{sec:contrast}, and \ref{sec:seed} show each of these choices changes the conclusion.

\paragraph{Prior computational tests of narrowing.}
Two lines bear most directly on ours. \citet{schatz2021} train models on realistic
child-centred audio and reproduce the native-language advantage only partially and without a clean
narrowing decline, arguing that phonetic \emph{categories} need not be learned at all---perception
may reorganise as a continuous space. \citet{millet2022} compare CPC, wav2vec2, and HuBERT against
French- and English-listener perceptual spaces and report a small native-language effect for CPC
but a largely \emph{language-universal} space for the masked-prediction models. Both attach the
objective to a model rather than isolating it (one model per objective, full scale,
whole-model or final-layer geometry), and---read across the two objective families---their
conclusions about which objective specializes do not obviously line up. We make the objective the
sole manipulated variable (same encoder, data, and seed) and read discriminability \emph{change}
layer by layer. This is a different observable from their static perceptual-space geometry: a model
can raise non-native ABX while still warping its space toward the native language, so our
transfer-direction result and their geometry results are complementary rather than
directly commensurable.

\section{Methods}

\paragraph{Model.}
Single-encoder Transformer: log-mel \(\rightarrow\) \texttt{Linear(80,384)} \(\rightarrow\)
4\,\(\times\)\,\texttt{TransformerEncoderLayer(d=384, heads=6, ff=1536)}. 7.1\,M parameters
(\texttt{baseline\_mini}). A brain-mapped \texttt{DualCodeModel} (\S\ref{sec:dualcode}) shares the
encoder scale and adds continuum-memory-system (CMS; \citealp{behrouz2025}) frequency-gated regions
mapped to STG/LIFG/dorsal/MTG.

\paragraph{Data.}
Child-directed speech: Providence (CHILDES; \citealp{demuth2006}), 176.7\,h. Read speech:
ZeroSpeech 2017 English/French/Mandarin 1\,s clips. All runs train 10\,k steps at batch 4, 1\,s
crops \(=\) 11.1\,h of exposure---\(4.5\times\) below the \(\sim\)50\,h at which large-scale
simulations first report robust phonemic ABX \citep{lavechin2025}.

\paragraph{Objectives.}
Selected by one config key so the encoder, data pipeline, and seeding are shared and only the
top-side supervision differs. Reconstruction: mask 15\% of mel frames, predict the original, MSE
on masked positions. Prediction: bilinear next-frame InfoNCE, temperature 0.1.

\paragraph{Evaluation.}
Phoneme ABX \citep{schatz2013}, 5000 within-context cross-speaker triplets per arm
(English/French/Mandarin), DTW-aggregated per-frame cosine. Per-triplet scores are stored so
contrast-level decomposition and uncertainty are recoverable without retraining.

\paragraph{Statistics.}
Ten seeds for the objective comparison and the crossover; three for the geometry and dual-code
probes; single-seed gates (confirmed at additional seeds where cheap) for the divergence search.
The \textbf{seed is the unit of replication}: trajectory changes are aggregated across seeds with a
\(t\) interval; within-seed uncertainty uses a paired bootstrap over triplets. One training run per
seed yields all milestones \(\{0,1\mathrm{k},5\mathrm{k},10\mathrm{k}\}\) via checkpoints.
Reported \(p\)-values are per-test and uncorrected: the first-order objective contrast
(Table~\ref{tab:contrast}) survives any multiple-comparison correction by orders of magnitude,
while the borderline tests (\S\ref{sec:contrast} contrast decomposition, \S\ref{sec:seed} gap
inversion) should be read as nominal.

\paragraph{Crossover.}
Train reconstruction on each of English/French/Mandarin read speech (one register), evaluate all
three arms.

\section{Results}

\subsection{The objective sets the direction of transfer}
\label{sec:direction}

\begin{figure}[t]
\centering
\includegraphics[width=\linewidth]{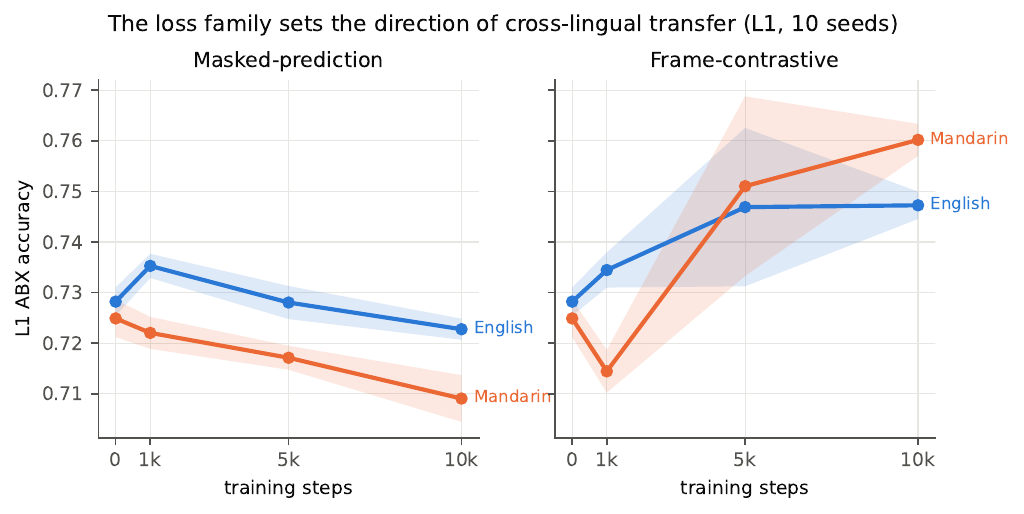}
\caption{The learning objective sets the direction of cross-lingual transfer (L1, ten seeds).
Masked-prediction (left): the native--non-native gap grows via the Mandarin arm declining.
Frame-contrastive (right): Mandarin crosses above English---the gap inverts. Bands are seed-level
intervals.}
\label{fig:direction}
\end{figure}

Encoder, data, and seed identical; only the objective varies. At L1, over 10\,k steps, native
(English) \(\Delta = \mnus0.006\) and non-native (Mandarin) \(\Delta = \mnus0.016\) under
reconstruction, versus \(+0.019\) and \(+0.035\) under prediction. The same-loss contrast on the
Mandarin arm---how much more it gains under prediction than reconstruction---is the paper's
strongest result (Table~\ref{tab:contrast}), significant at every layer and unanimous in sign
across all ten seeds at every layer for both objectives (40 of 40 layer-condition cells). The
objective flips the direction of cross-lingual transfer without exception (Fig.~\ref{fig:direction}).
Neither objective reproduces the developmental signature: reconstruction gets the non-native
decline without native gain; prediction improves both arms, Mandarin more, from a lower start.

\begin{table}[t]
\centering
\small
\caption{The objective contrast (prediction \(-\) reconstruction) on the Mandarin arm, by layer
(ten seeds, seed-level 95\% CI).}
\label{tab:contrast}
\begin{tabular}{llll}
\toprule
layer & contrast & 95\% CI & \(p\) \\
\midrule
L1 & \(+0.051\) & \([+0.046,+0.056]\) & \(3\times10^{-8}\) \\
L2 & \(+0.055\) & \([+0.047,+0.063]\) & \(<10^{-6}\) \\
L3 & \(+0.062\) & \([+0.052,+0.072]\) & \(<10^{-6}\) \\
L4 & \(+0.048\) & \([+0.037,+0.058]\) & \(<10^{-6}\) \\
\bottomrule
\end{tabular}
\end{table}

\subsection{The decline is L1--L2 and the final layer inverts it}
\label{sec:layer}

\begin{figure}[t]
\centering
\includegraphics[width=0.6\linewidth]{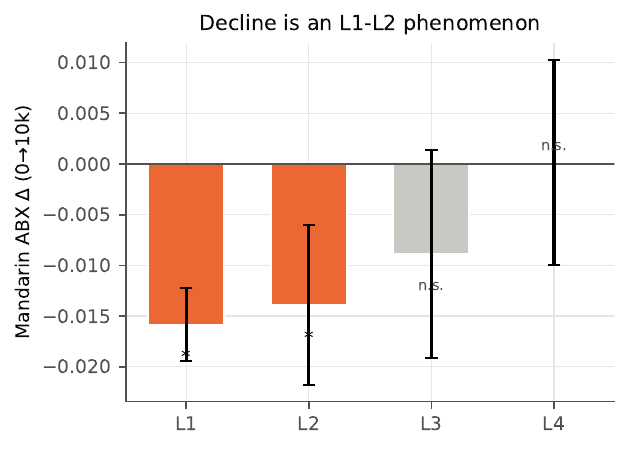}
\caption{The masked-prediction Mandarin decline is significant at L1--L2 (\(*\)) and gone by
L3--L4 (n.s.). Bars: seed-level mean \(\Delta\); whiskers: 95\% CI.}
\label{fig:layerwise}
\end{figure}

Reconstruction Mandarin \(\Delta\) by layer: L1 \(\mnus0.016\) (\(p=2\times10^{-6}\)), L2
\(\mnus0.014\) (\(p=0.003\)), L3 \(\mnus0.009\) (n.s.), L4 \(+0.000\) (n.s.). The decline is a
first-half-of-network phenomenon, gone by L3 (Fig.~\ref{fig:layerwise}). At L1 the gap grows
because the non-native arm \emph{declines}; at L4 a gap of similar size grows through parallel
drift with no decline. The field's final-layer ABX measures the second mechanism and misses the
first.

\subsection{At the contrast level, the exclusively-non-native contrasts decline most}
\label{sec:contrast}

Decomposing the all-contrasts Mandarin arm by whether a contrast exists in English (a linguistic
English-absent set: tonal, retroflex, alveolo-palatal, aspirated stops), under reconstruction at
L1: English-shared \(\mnus0.014\), English-absent \(\mnus0.046\) (difference \(\mnus0.032\),
\(p=0.0017\)). This is not a floor effect---English-absent starts \emph{lower} (0.65 vs 0.73) yet
declines more. The contrasts that decline most are precisely those the infant paradigm tests.

\subsection{The crossover: two mechanisms, not one}
\label{sec:crossover}

\begin{figure}[t]
\centering
\begin{minipage}{0.48\linewidth}
\centering
\includegraphics[width=\linewidth]{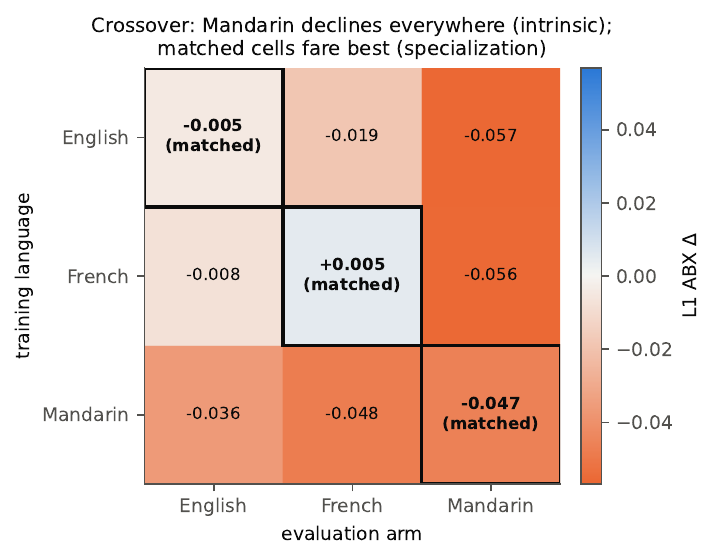}
\caption{Training-language crossover (L1 \(\Delta\)). The Mandarin column is uniformly negative
(arm-intrinsic difficulty); matched-diagonal cells (boxed) fare best in their column
(specialization).}
\label{fig:crossover}
\end{minipage}\hfill
\begin{minipage}{0.48\linewidth}
\centering
\includegraphics[width=\linewidth]{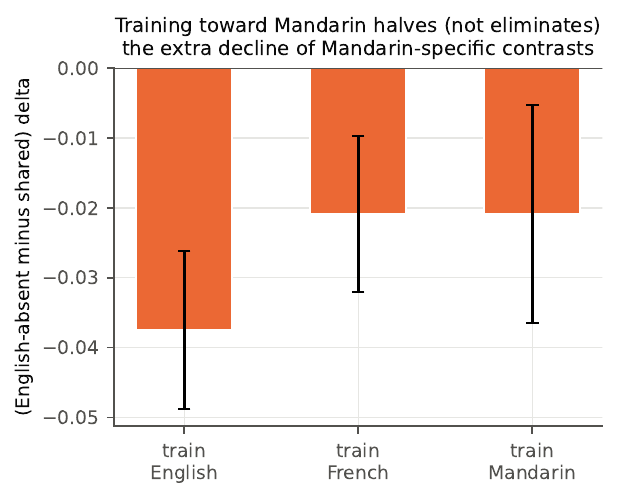}
\caption{The extra decline of Mandarin-specific (English-absent) contrasts by training language:
steepest under English, roughly halved but not eliminated under Mandarin training.}
\label{fig:gradient}
\end{minipage}
\end{figure}

Training reconstruction on each language and evaluating all three arms (L1 \(\Delta\), ten seeds)
gives the matrix in Fig.~\ref{fig:crossover}. \textbf{Arm-intrinsic difficulty (large):} the
Mandarin arm declines under every training language, including Mandarin (\(\mnus0.047\)); pooled
column means English \(\mnus0.015\), French \(\mnus0.016\), Mandarin \(\mnus0.051\).
\textbf{Language specialization (smaller, significant):} pooled matched-vs-mismatched \(+0.022\),
significant at all four layers (\(p=10^{-4}\) at L1--L3, \(p=10^{-3}\) at L4); five of six per-pair
contrasts individually significant. \textbf{The decisive cell:} training on Mandarin does not
rescue the Mandarin arm---it still falls \(\mnus0.047\). The pattern is neither pure degradation
(specialization is significant) nor pure specialization (the hardest arm is not rescued):
reconstruction produces both. At the contrast level (Fig.~\ref{fig:gradient}) the extra decline of
Mandarin-specific contrasts halves as training moves toward Mandarin (\(\mnus0.038 \rightarrow
\mnus0.021\)) but does not vanish.

\subsection{Reconstruction falls below its input floor; prediction rises above it}
\label{sec:geometry}

Against the raw-mel ABX floor (no encoder), which is language-symmetric (native 0.733, Mandarin
0.731---no input-level English advantage), L1 ABX at 10\,k: reconstruction native 0.720
(\(\mnus0.013\) vs floor), Mandarin 0.708 (\(\mnus0.023\)); prediction native 0.742 (\(+0.009\)),
Mandarin 0.758 (\(+0.026\)). Reconstruction pushes L1 below the discriminability already in its
input, more for non-native; prediction pushes it above. Effective rank \emph{increases} under both
(reconstruction \(+4.6/+4.7\), prediction \(+8.5/+7.4\)), so this is not dimensional collapse but a
reallocation of capacity. The floor makes ``degradation'' and ``improvement'' absolute statements,
and gives \S\ref{sec:direction} its mechanism.

\subsection{The magnitude is register-sensitive}
\label{sec:register}

Training reconstruction on read speech vs child-directed speech (English, both ten seeds), the
Mandarin decline is \(3.6\times\) steeper on read (L1 \(\mnus0.057\) vs \(\mnus0.016\), difference
\(\mnus0.041\), \(p<10^{-4}\)), while the native arm is register-robust (agreement within 0.008).
The sign is register-invariant; the magnitude is not. This explains why a child-directed-only
pilot saw a weak decline: CDS minimizes it.

\subsection{The three-seed budget cannot see these effects reliably}
\label{sec:seed}

\begin{figure}[t]
\centering
\includegraphics[width=0.62\linewidth]{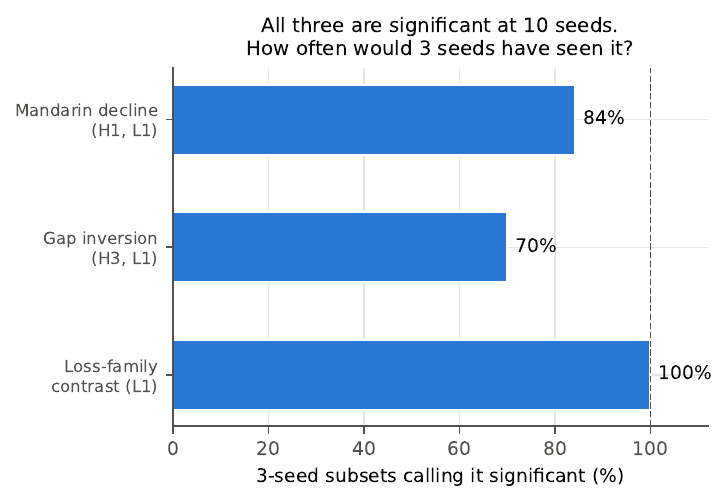}
\caption{All three claims are significant at ten seeds; the bars show the fraction of three-seed
subsets that would also call each significant. The gap inversion---significant at \(n=10\)---is
seen by only 70\% of three-seed draws.}
\label{fig:fragility}
\end{figure}

Enumerating all \(\binom{10}{3}=120\) three-seed subsets and re-running the seed-level test: the
loss-family contrast is called significant by 100\% of subsets, the H1 Mandarin decline by 84\%,
but the gap inversion---also unambiguous at ten seeds (\(p=8\times10^{-6}\))---by only 70\%, subset
\(p\)-values spanning 0.000--0.209 (Fig.~\ref{fig:fragility}). Our own original seed 0--2 sample
landed at \(p=0.073\), in the 30\% that miss. The field's customary seed budget under-powers
exactly the developmental effects it searches for.

\subsection{A brain-circuit dual-code architecture reproduces narrowing at no region}
\label{sec:dualcode}

Probing all four Yoo (2013)-mapped regions of a CMS-gated DualCodeModel (three seeds,
reconstruction): no region shows the native-favoring narrowing gap (every gap CI spans zero). But
the regions are not static---the fast LIFG (articulatory) \emph{improves} (\(+0.07\) to \(+0.09\))
while the slow MTG (acoustic-phonetic) \emph{collapses} (\(\mnus0.16\) to \(\mnus0.18\)). The
frequency gating produces a strong articulatory-up / acoustic-phonetic-down dissociation, not a
language specialization. The dissertation maps MTG to the long-term store, yet under reconstruction
it degrades most---the slow region lags rather than consolidates, consistent with
\S\ref{sec:geometry}. So the architecture is far from inert: it produces a strong
articulatory-up / acoustic-phonetic-down dissociation, but that dissociation is not
language-selective: the \emph{narrowing}-shaped, native-favouring effect is carried by the
objective, not the region wiring.

\subsection{The divergence search: six configurations, none reproduce the signature}
\label{sec:divergence}

\begin{figure}[t]
\centering
\includegraphics[width=\linewidth]{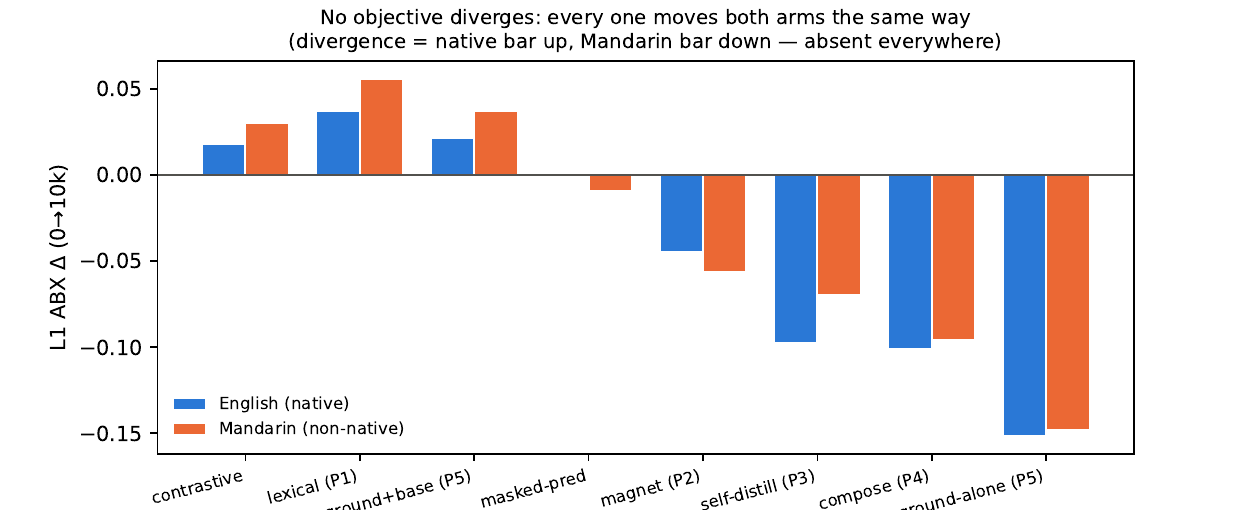}
\caption{No objective diverges: every configuration moves both arms the same way. Divergence would
be a native (blue) bar up beside a Mandarin (orange) bar down in the same group---absent
everywhere.}
\label{fig:map}
\end{figure}

Results \S\ref{sec:direction}--\ref{sec:dualcode} show the objective sets the \emph{direction} but
no objective produces the full developmental \emph{signature}---native improves \emph{and}
non-native declines (divergence). We searched for it directly with six objective configurations,
each engineered to produce divergence and grounded in a distinct account of narrowing
(Table~\ref{tab:map}, Fig.~\ref{fig:map}).

\emph{Lexical top-down} \citep{feldman2013}: a word-supervised-contrastive auxiliary lifts
native-word contrasts---but word discrimination is itself a discriminative objective, so it lifts
both arms language-agnostically. \emph{Perceptual magnet} \citep{kuhl1992,maye2002}: a
vector-quantization codebook \citep{oord2017} of native prototypes---but frame-level quantization
sinks both arms by collapsing sub-phonemic detail. \emph{Self-distillation}
\citep{baevski2022,caron2021,mcclelland1995}: a data2vec EMA teacher supplies a native-consolidated
target---but the reconstruction-like teacher-prediction degrades native most. \emph{Composition}:
lexical lift plus prototype compression cracks native only transiently before the compression
drags it down. \emph{Semantic grounding}: regress each native word's audio onto its GloVe embedding
\citep{pennington2014}---with a base it joins the both-up family; alone it sinks both by collapsing
phonetic detail to word-meaning, which is coarser than phonetics.

A single objective moves both languages the \emph{same} way because it shapes a representation
\emph{shared} between them (the \S\ref{sec:geometry} floor result in another guise). Sharpening and
word-grounding lift both; compression, consolidation, and grounding-alone sink both; composing a
lift with a non-selective compression gives a non-selective net. The signature requires
\textbf{selectivity in the compression itself}: to spare native, the objective must know which
contrasts are native-relevant---a signal finer than any single loss or word-level target we tested
carries.

\begin{table}[t]
\centering
\small
\caption{The divergence-search mechanism map: six configurations, L1 \(\Delta\) (seed-0 gates,
confirmed at further seeds where cheap). None diverges (native up \emph{and} non-native down).}
\label{tab:map}
\begin{tabular}{lllc}
\toprule
objective & mechanism & native / non-native \(\Delta\) & shape \\
\midrule
lexical & sharpening & \(+0.037 / +0.056\) & both up \\
grounding \(+\) base & grounding & \(+0.021 / +0.037\) & both up \\
masked-pred & reconstruction & \(\mnus0.006 / \mnus0.016\) & both down \\
magnet & compression & \(\mnus0.045 / \mnus0.056\) & both down \\
self-distill & consolidation & \(\mnus0.098 / \mnus0.070\) & both down (native more) \\
lexical \(+\) magnet & composition & \(\mnus0.101 / \mnus0.096\) & both down \\
grounding alone & grounding & \(\mnus0.152 / \mnus0.148\) & both down \\
\bottomrule
\end{tabular}
\end{table}

\section{Discussion}

\paragraph{What produces narrowing: the objective.}
The results converge on one claim: the learning objective is the first-order determinant of
narrowing-shaped representation change. Reconstruction produces contrast-specific,
training-language-specific, layer-localized loss of non-native discriminability that falls below
the input-feature floor; prediction produces the opposite; the brain-mapped architecture (the dual
code) does not by itself carry the language-selective effect---it yields an articulatory/acoustic
dissociation instead. For developmental neuroscience this is a testable dissociation: if early auditory
learning approximates a reconstruction-like objective, narrowing-shaped loss should follow---most
where the input register is most canonical (\S\ref{sec:register}); if a prediction-like objective,
it should not. The dual-stream mapping suggests a concrete hypothesis: the ventral
acoustic-phonetic stream (MTG) and the dorsal articulatory stream (LIFG) may implement different
objectives, and \S\ref{sec:dualcode}'s dissociation under a single objective is a first, indirect
probe.

\paragraph{The selectivity requirement, and the dorsal buffer.}
The divergence search turns a series of negatives into a positive statement: divergence needs
\emph{selectivity}---the force that lifts native must be distinct from the force that touches
non-native, which a single loss on a shared representation cannot supply. This is not a scale
problem (prediction improves both arms \emph{more} with more data, never diverging); it is
structural. This is, we argue, the computational content of the Yoo (2013) dorsal buffer, whose
role is semantic association---binding sound to meaning, a native-relevance signal external to the
acoustic objective. Our tests sharpen \emph{what} that signal must be: not a word-contrastive push
(language-agnostic) nor a word-meaning regression (coarser than phonetics), but something that
reorganises phonetic categories toward native ones while sparing them---subtler than any single
loss or word-level target buildable at this scale.

\paragraph{What we do not claim.}
No objective reproduces the full signature at 11.1\,h; the objective is necessary-looking but not
sufficient. The gap inversion is significant at ten seeds but reported as secondary: it is the
claim most sensitive to seed draw. ``Prediction'' here is a simplified CPC without an
autoregressive aggregator. The divergence-search gates are single- to few-seed with large,
mechanistically distinct directions.

\paragraph{Limitations, and what survives them.}
Four limitations bound the study, and it is worth being explicit about which touch the central
claim and which touch only its scope. \emph{Scale}---11.1\,h/seed, \(4.5\times\) below the
\(\sim\)50\,h at which large-scale simulations first report robust phonemic ABX---cannot confound
the first-order result by construction, because the objective contrast is a same-scale, same-data,
same-seed difference in which scale cancels; it could overturn the conclusion only if the
\emph{sign} of the contrast reversed with more data, and the geometry result (\S\ref{sec:geometry},
a reallocation of capacity rather than an undertraining artifact) gives no reason to expect that.
Scale therefore bounds the already-negative \emph{signature} claim, not the \emph{direction} claim.
\emph{Register entangled with corpus} (Providence child-directed vs ZeroSpeech read) limits only the
interpretation of the \(3.6\times\) magnitude effect: the \emph{sign} is register- and
corpus-invariant (\S\ref{sec:register}), so the direction result holds within each corpus
separately. \emph{Three-seed geometry and dual-code probes} are supporting rather than
load-bearing---the direction and crossover results are ten-seed, and \S\ref{sec:seed} directly
measures that the load-bearing contrast survives \(100\%\) of three-seed subsets---with one honest
exception: the dual-code ``no region narrows'' finding is a null at \(n=3\), the single place our
defense rests on an argument (a \(\sim\)0.02 language-selective gap would surface in the majority of
three-seed draws) rather than on measurement. \emph{Single architecture and language family} bounds
generality, but the effect already reproduces across two architectures (baseline and dual-code) and
on a typologically distant, tonal non-native target, and its mechanism---the input-feature
floor---is architecture-agnostic in principle. In sum, the direction result is immune to scale by
design, robust to register and seed budget by measurement, and reproduced across two architectures;
the one residual that remains an argument rather than a result is the three-seed dual-code null.

\paragraph{Future work.}
These limitations convert cleanly into three experiments, each of which would turn a surviving
argument into evidence: a ten-seed re-run of the geometry and dual-code probes, to power the one
null the defense currently rests on; a scale ladder (11/45/176\,h) confirming the contrast keeps its
sign and that prediction's both-arms gain \emph{grows} rather than diverges; and a corpus-matched
register contrast de-confounding the magnitude effect. Beyond closing limitations, the substantive
open lever for the full signature is a second, native-relevance signal richer than a word
label---utterance-level visual grounding is its most natural form (the dorsal-buffer signal as
such)---since scale alone does not diverge and would have to be paired with a selective or grounded
objective. Per-phoneme-category geometry would localize where in the space the reallocation falls.

\section{Conclusion}

Perceptual narrowing is a developmental-neuroscience phenomenon in search of a computational cause.
We show, in a controlled ten-seed self-supervised setting, that the learning objective is that
cause at first order: reconstruction degrades non-native phoneme discrimination---contrast-,
training-language-, and layer-specifically, below the representation's own input floor---while
prediction improves it, a same-encoder contrast of \(+0.051\) unanimous across twenty runs. The
decline is two mechanisms, its magnitude is register-sensitive, and a brain-circuit-mapped
architecture reproduces none of the language-selective narrowing, yielding an articulatory/acoustic
dissociation instead. An effect unambiguous at ten seeds is missed by 30\% of three-seed
studies. And a direct search over six objective configurations shows that none reproduces the full
signature, because an objective on a shared representation moves both languages together; the
signature requires a selectivity finer than a word label, which our tests bound to utterance-level
grounding or a mechanism subtler than any single loss. The objective sets the direction of
narrowing; the second, native-relevance signal is what would turn direction into the developmental
signature, and that, not the architecture or the scale alone, is where the developmental question
should be asked next.

\paragraph{Reproducibility.}
Every result derives from public corpora and a fully specified setup. \emph{Data:} the Providence
corpus of child-directed speech (CHILDES; 176.7\,h) and the ZeroSpeech 2017 English/French/Mandarin
read-speech sets, with GloVe 6B (50d) word vectors for the semantic-grounding probe---all publicly
available. \emph{Model:} a 7.1\,M-parameter single-encoder Transformer (log-mel \(\rightarrow\)
\texttt{Linear(80,384)} \(\rightarrow\) 4\,\(\times\)\,\texttt{TransformerEncoderLayer}(\(d\)=384,
6 heads, feed-forward 1536)) and, for the architectural probe, a CMS frequency-gated dual-code
variant of matched scale. \emph{Setup:} PyTorch on Apple-silicon (MPS); per seed, 10\,k optimiser
steps at batch 4 over 1\,s crops (11.1\,h of exposure), with fixed seeds and a single run per seed
yielding all milestone checkpoints \(\{0,1\mathrm{k},5\mathrm{k},10\mathrm{k}\}\). \emph{Evaluation:}
phoneme ABX over 5000 within-context, cross-speaker triplets per arm, scored by DTW-aggregated
per-frame cosine, with per-triplet scores retained so that every contrast-level decomposition and
uncertainty estimate is recomputable without retraining.

\end{document}